\documentclass[10pt,towclome]{bmc_article}

\usepackage{amssymb}
\usepackage{amsmath}
\usepackage{algorithmic}
\usepackage{algorithm}
\usepackage{graphicx}
\usepackage{subfigure}
\usepackage{array}
\usepackage{multirow,tabularx}

\usepackage[pagewise]{lineno}
\usepackage{cite}
\usepackage{url}
\usepackage{ifthen}
\usepackage{multicol}
\usepackage[utf8]{inputenc}
\usepackage{pdfsync}
\usepackage{color}

\newboolean{publ}

\newenvironment{bmcformat}{\baselineskip20pt\sloppy\setboolean{publ}{false}}{\baselineskip20pt\sloppy}

\begin{document}
\begin{bmcformat}

\title{Multiple graph regularized protein domain ranking}

\author{
Jim Jing-Yan Wang$^{1}$,
\email{Jim Jing-Yan Wang - jimjywang@gmail.com}
Halima Bensmail$^2$ and
\email{Halima Bensmail - hbensmail@qf.org.qa}
Xin Gao\correspondingauthor$^{1,3}$
\email{Xin Gao - xin.gao@kaust.edu.sa}
}

\address{
\iid(1)
Computer, Electrical and Mathematical Sciences and Engineering Division,
King Abdullah University of Science and Technology (KAUST),
Thuwal, 23955-6900, Saudi Arabia\\
\iid(2)
Qatar Computing Research Institute, Doha 5825, Qatar\\
\iid(3)
Computational Bioscience Research Center, King Abdullah University of
Science and Technology (KAUST), Thuwal 23955-6900, Saudi Arabia
}

\maketitle

\begin{abstract}
\begin{description}
\item[Background] Protein domain ranking is a fundamental task in structural biology. Most protein domain ranking methods rely on the pairwise comparison of protein domains while neglecting the global manifold structure of the protein domain database. Recently, graph regularized ranking that exploits the global structure of the graph defined by the pairwise similarities has been proposed. However, the existing graph regularized ranking methods are very sensitive to the choice of the graph model and parameters, and this remains a difficult problem for most of the protein domain ranking methods.
\item[Results] To tackle this problem, we have developed the Multiple Graph regularized Ranking algorithm, MultiG-Rank. Instead of using a single graph to regularize the ranking scores, MultiG-Rank approximates the intrinsic manifold of protein domain distribution by combining multiple initial graphs for the regularization. Graph weights are learned with ranking scores jointly and automatically, by alternately minimizing an objective function in an iterative algorithm. Experimental results on a subset of the ASTRAL SCOP protein domain database demonstrate that MultiG-Rank
achieves a better ranking performance than single graph regularized ranking methods and pairwise similarity based ranking methods.
\item[Conclusion] The problem of graph model and parameter selection in graph regularized protein domain ranking can be solved effectively by combining multiple graphs. This aspect of generalization introduces a new frontier in applying multiple graphs to solving protein domain ranking applications.
\end{description}
\end{abstract}

\ifthenelse{\boolean{publ}}{\begin{multicols}{2}}{}

\newpage

\section*{Background}

Proteins contain one or more domains each of which
could have evolved independently from the rest of the protein structure
and which could have unique functions \cite{Domain2011a,Domain2011b}.
Because of molecular evolution,
proteins with similar sequences often share similar folds and structures.
 Retrieving and ranking protein domains that are similar to a query protein domain from a protein domain database
are critical
tasks for the analysis of
protein structure, function, and evolution
 \cite{Zhang2010,Stivala2009,Stivala2010}.
The similar protein domains that are classified by a ranking system may help researchers infer the functional properties of a query domain from the functions of the returned protein domains.

The output of a
ranking procedure is usually a list of database protein domains
that are ranked in descending order according to a measure of their similarity to the query domain.
The choice of a similarity measure largely defines the performance of a ranking system as argued previously \cite{BaiGT2010}.
A large number of algorithms for computing similarity as a ranking score have been developed:
\begin{description}
\item[Pairwise protein domain comparison algorithms] compute the similarity between a pair of protein domains either by protein domain structure alignment or
by comparing protein domain features.
\emph{Protein structure alignment based methods} compare protein domain structures at the level of residues and sometime even atoms, to detect structural
similarities with high sensitivity and accuracy.
For example,
Carpentier et al. proposed YAKUSA \cite{YAKUSA2005}
which compares protein  structures using one-dimensional
characterizations based on protein  backbone
internal angles,
while
Jung and Lee proposed SHEBA  \cite{SHEBA1990}
for
structural database scanning based on environmental profiles.
\emph{Protein domain feature based methods} extract structural features from protein domains and compute their similarity using a similarity
or distance function.
For example, Zhang et al.
used the 32-D tableau feature vector in a comparison procedure called IR tableau
\cite{Zhang2010}, while Lee and Lee
introduced a measure called WDAC (Weighted Domain Architecture Comparison) that is used in the protein domain comparison context \cite{Cosine2009}. Both these methods use cosine similarity for comparison purposes.

\item[Graph-based similarity learning algorithms]
use the traditional protein domain comparison methods mentioned above
that focus on detecting pairwise sequence
alignments while
neglecting all other protein domains in the database and their distributions.
To tackle this problem,
a graph-based transductive similarity learning algorithm
has been proposed \cite{BaiGT2010,Weston2006}.
Instead
of computing pairwise similarities for protein domains,
graph-based methods take advantage
of the graph formed by the existing protein domains.
By propagating similarity measures between the query protein domain and the database protein domains via
graph transduction (GT),  a better metric for ranking database protein domains can be learned.
\end{description}

The main component of graph-based ranking is the construction of a graph  as the estimation of
intrinsic
manifold of the database.
As argued by Cai et al. \cite{Cai2011},
there are many ways to define different graphs with different models and parameters.
However, up to now,
there are, in general, no explicit rules for choice of graph models  and parameters.
In \cite{BaiGT2010}, the graph parameters were determined by a grid-search of different pairs of parameters.
In \cite{Cai2011}, several graph models were considered for graph regularization, and exhaustive experiments were carried out for the selection of a graph model and its parameters .
However, these kinds of grid-search strategies select  parameters from discrete values in the
parameter space, and thus lack the ability to approximate an optimal solution.
At the same time, cross-validation \cite{CrossValidation2006,CrossValidation2009} can be used for parameter selection, but it
does not always scale up very well for many of the graph parameters, and sometimes it might
over-fit the training
and validation set while not generalizing well on the query set.

In \cite{Geng2012}, Geng et al. proposed an ensemble manifold
regularization (EMR) framework that combines the automatic intrinsic manifold approximation
and semi-supervised learning (SSL) \cite{SSL2010,SSL2011} of a support vector machine (SVM) \cite{SVM2011a,SVM2011b}.
Based on the EMR idea, we attempted to solve the problem of graph model and parameter selection
by fusing multiple graphs to obtain a ranking score learning framework for protein domain ranking.
We first outlined the graph regularized ranking score learning framework
by optimizing ranking score learning with both relevant and graph constraints , and then
generalized it to the multiple graph case.
First a pool of initial guesses of the graph Laplacian with different graph models and parameters is computed,
and then they are combined linearly to approximate the intrinsic manifold.
The optimal graph model(s) with optimal parameters is selected by assigning larger weights to them.
Meanwhile, ranking score learning is also restricted to be smooth
along the estimated graph.
 Because the graph weights and ranking scores are learned
jointly, a unified
objective function is obtained.
The objective function is optimized alternately and conditionally with respect to multiple graph weights
and ranking scores in an iterative algorithm.
We have named our \textbf{Multi}ple \textbf{G}raph regularized \textbf{Rank}ing method \textbf{MultiG-Rank}. It is composed of an off-line graph weights learning algorithm and an on-line ranking algorithm.

\section*{Methods}
Graph model and parameter selection
Given a data set of protein domains represented by their tableau 32-D feature vectors \cite{Zhang2010} $\mathcal{X} = \{x_1, x_2, \cdots, x_N\}$,
where $x_i\in \mathbb{R}^{32}$ is the tableau feature vector of $i$-th protein domain, $x_q$  is the query protein domain, and the others are database protein domains.
We define the ranking score vector as $\textbf{f} = [f_1, f_2, ..., f_N]^\top \in  \mathbb{R}^N$ in which $f_i$ is the
ranking score of $x_i$ to  the query domain.
The problem is to rank the protein domains in $\mathcal{X}$ in descending order according to their
ranking scores and return several of the top ranked domains as the ranking results so that the returned protein domains are as relevant to the query as possible.
Here we define two types of protein domains: \emph{relevant} when they belong to the same SCOP fold type \cite{SCOP2009}, and \emph{irrelevant} when they do not.
We denote the SCOP-fold labels of protein domains in $\mathcal{X}$ as $\mathcal{L} = \{l_1, l_2, ..., l_N\}$, where
$l_i$ is the label of $i$-th protein domain and $l_q$ is the query label.
The optimal ranking scores of relevant protein domains $\{x_i\}, l_i=l_q$ should be larger than the irrelevant ones
$\{x_i\}, l_i\neq l_q$, so that the
relevant protein domains will be returned to the user.

\subsection*{Graph regularized protein domain ranking}

We applied
two constraints on the optimal ranking score vector $\textbf{f}$ to learn the optimal ranking scores:

\begin{description}
\item[Relevance constraint]
Because the query protein domain reflects the search intention of the user, $f$ should be consistent with protein domains that are relevant to the query
.
We also define a relevance vector of the protein domain as $\textbf{y} = [y_1, y_2, \cdots, y_N]^\top \in \{1,0\}^N$ where
$y_i = 1$, if $x_i$ is relevant to the query and $y_i = 0$ if it is not.
Because the type label $l_q$ of a query protein domain $x_q$ is usually unknown,
we know only that the query is relevant to itself and
have no prior knowledge of whether or not others are relevant;
therefore, we can only set $y_q=1$ while $y_i,~i\neq q$ is unknown.

To assign different weights to different protein domains in $\mathcal{X}$,
we define a diagonal matrix $U$
as $U_{ii}=1$ when $y_i$ is known, otherwise $U_{ii}=0$.
To impose the relevant constraint to the learning of $f$,
we aim to minimize the following
objective function:

\begin{equation}
\label{equ:Qr}
\begin{aligned}
\underset{\textbf{f}}{min}~ O^r(\textbf{f})
&=\sum_{i=1}^N (f_i-y_i)^2 U_{ii}\\
&=(\textbf{f} - \textbf{y})^\top U(\textbf{f} - \textbf{y})
\end{aligned}
\end{equation}

\item[Graph constraint]
$f$ should also be consistent with the local distribution found in the protein domain database.
The local distribution was embedded into a $K$ nearest neighbor graph $\mathcal{G}=\{\mathcal{V},\mathcal{E},W\}$.
For each protein domain $x_i$, its $K$ nearest neighbors, excluding itself, are denoted by $\mathcal{N}_i$.
The node set $\mathcal{V}$ corresponds to $N$ protein domains in $\mathcal{X}$, while
$\mathcal{E}$ is the edge set, and
$(i,j)\in \mathcal{E}$ if $x_j\in \mathcal{N}_i$ or $x_i\in \mathcal{N}_j$.
{
The weight of an edge $(i,j)$ is denoted as $W_{ij}$ which can be computed using different graph definitions and parameters as described in the next section.
The edge weights are further organized in a weight matrix $W = [W_{ij}] \in \mathbb{R}^{N\times N}$},
where $W_{ij}$
is the weight of edge {$(i,j)$}.
We expect that if two {protein domains}
$x_i$ and $x_j$ are close (i.e.,$W_{ij}$ is big), then $f_i$ and $f_j$ should also
be close.
To impose the graph constraint to the learning of $f$,
we aim to minimize the following
objective function:

\begin{equation}
\label{equ:Og}
\begin{aligned}
\underset{\textbf{f}}{min}~ O^g(f)
&=\frac{1}{2}\sum_{i,j=1}^N (f_i-f_j )^2 W_{ij}\\
&=\textbf{f}^\top D \textbf{f} - \textbf{f}^\top W \textbf{f}\\
&=\textbf{f}^\top L \textbf{f}
\end{aligned}
\end{equation}
where $D$ is a
diagonal matrix whose entries are $D_{ii}=\sum_{i=1}^N W_{ij}$ and $L = D -W$ is the
graph Laplacian matrix.
This is a basic identity in spectral graph theory and it provides some
insight into the remarkable properties of the graph Laplacian.
\end{description}

When the two constraints are combined,
the learning of $\textbf{f}$ is based on the minimization of the
following objective
function:

\begin{equation}
\label{equ:Of}
\begin{aligned}
\underset{\textbf{f}}{min}~ O(\textbf{f})&=O^r(\textbf{f}) + \alpha O^g(\textbf{f})\\
&=(\textbf{f} - \textbf{y})^\top U(\textbf{f} - \textbf{y}) + \alpha \textbf{f}^\top L \textbf{f}
\end{aligned}
\end{equation}
where $\alpha$ is a trade-off parameter of the smoothness penalty.
The solution is obtained by setting the derivative of $O(\textbf{f})$ with respect to $\textbf{f}$ to zero as
$\textbf{f}=(U+\alpha L)^{-1}U\textbf{y}$.
In this way, information from both
the query protein domain provided by the user and
the relationship of all the protein domains in $\mathcal{X}$
are used to
rank the protein domains in $\mathcal{X}$.
The query information is embedded in $y$ and $U$, while the protein domain relationship information is embedded in $L$.
The final ranking results are obtained by balancing the two
sources of information.
In this paper, we call this method \textbf{G}raph regularized \textbf{Rank}ing (G-Rank).

\subsection*{Multiple graph learning and ranking: MultiG-Rank}

Here we describe the multiple graph learning
method to directly learn a self-adaptive graph for ranking regularization The graph is assumed to be a linear
combination of multiple predefined graphs (referred
to as base graphs).
The graph weights are learned in a supervised way by considering the
SCOP fold types of the protein domains in the database.

\subsubsection *{Multiple graph regularization}

The main component of graph regularization is the construction of a graph.
As described previously, there are many ways to find the neighbors $\mathcal{N}_i$ of $x_i$ and to define the weight matrix $W$ on
the graph \cite{Cai2011}.
Several of them are as follows:
\begin{itemize}

\item
\textbf{Gaussian kernel weighted graph}: $\mathcal{N}_i$ of $x_i$ is found by comparing the squared Euclidean distance
as,

\begin{equation}
\begin{aligned}
||x_i-x_j||^2=x_i^\top x_i - 2 x_i^\top x_j +x_j^\top x_j
\end{aligned}
\end{equation}
and the weighting is computed using a Gaussian kernel as,

\begin{equation}
\begin{aligned}
W_{ij}=
\left\{\begin{matrix}
e^{-\frac{||x_i-x_j||^2}{2\sigma^2}}, &if~(i,j)\in \mathcal{E}\\
0, & else
\end{matrix}\right.
\end{aligned}
\end{equation}
where $\sigma$ is the bandwidth of the kernel.

\item
\textbf{Dot-product weighted graph}: $\mathcal{N}_i$ of $x_i$ is found by comparing the squared Euclidean distance
and the weighting is computed as the dot-product as,

\begin{equation}
\begin{aligned}
W_{ij}=
\left\{\begin{matrix}
x_i^\top x_j, &if~(i,j)\in \mathcal{E}\\
0, & else
\end{matrix}\right.
\end{aligned}
\end{equation}

\item \textbf{Cosine similarity weighted graph}:
$\mathcal{N}_i$ of $x_i$ is found by comparing
cosine similarity as,

\begin{equation}
\begin{aligned}
C(x_i,x_j)=\frac{x_i^\top x_j}{||x_i||||x_j||}
\end{aligned}
\end{equation}
and the weighting is also assigned as cosine similarity as,

\begin{equation}
\begin{aligned}
W_{ij}=
\left\{\begin{matrix}
C(x_i,x_j), &if~(i,j)\in \mathcal{E}\\
0, & else
\end{matrix}\right.
\end{aligned}
\end{equation}

\item \textbf{Jaccard index weighted graph}:
$\mathcal{N}_i$ of $x_i$ is found by comparing
the Jaccard index \cite{Jaccard2011} as,

\begin{equation}
\begin{aligned}
J(x_i,x_j)=\frac{|x_i\bigcap x_j|}{|x_i\bigcup x_j|}
\end{aligned}
\end{equation}
and the weighting is assigned as,

\begin{equation}
\begin{aligned}
W_{ij}=
\left\{\begin{matrix}
J(x_i,x_j), &if~(i,j)\in \mathcal{E}\\
0, & else
\end{matrix}\right.
\end{aligned}
\end{equation}

\item \textbf{Tanimoto coefficient weighted graph}:
$\mathcal{N}_i$ of $x_i$ is found by comparing
the Tanimoto coefficient as,

\begin{equation}
\begin{aligned}
T(x_i,x_j)=\frac{x_i^\top x_j}{||x_i||^2+||x_j||^2-x_i^\top x_j}
\end{aligned}
\end{equation}
and the weighting is assigned as,

\begin{equation}
\begin{aligned}
W_{ij}=
\left\{\begin{matrix}
T(x_i,x_j), &if~(i,j)\in \mathcal{E}\\
0, & else
\end{matrix}\right.
\end{aligned}
\end{equation}

\end{itemize}

With so many possible choices of graphs,
the
most suitable graph with its parameters for the protein domain ranking task is often not known
in advance;
thus, an exhaustive search on a predefined
pool of graphs is necessary.
When the size
of the pool becomes large, an
exhaustive search will be quite time-consuming and sometimes not possible.
Hence, a method for efficiently learning
an appropriate graph to make the performance
of the employed graph-based ranking method robust or
even improved is crucial for graph regularized ranking.
To tackle this problem we propose a multiple graph
regularized ranking framework,
that provides a series of initial guesses of the graph Laplacian
and combines them to approximate the intrinsic manifold in a conditionally
optimal way, inspired by a previously reported method \cite{Geng2012}.

Given a set of $M$ graph candidates $\{\mathcal{G}_1,\cdots,\mathcal{G}_M\}$,
we denote their corresponding candidate graph Laplacians as $\mathcal{T}=\{L_1,\cdots,L_M\}$.
By assuming that the optimal graph Laplacian
lies in the convex hull of the pre-given graph Laplacian candidates,
we constrain the search space of possible graph Laplacians o linear combination of $L_m$ in $\mathcal{T}$ as,

\begin{equation}
\label{equ:L}
\begin{aligned}
L=\sum_{m=1}^M \mu_m L_m
\end{aligned}
\end{equation}
where $\mu_m$ is the weight of $m$-th graph.
To avoid any negative contribution,
we further constrain
$\sum_{m=1}^M \mu_m=1,~\mu_m\geq 0.$

To use the information from
data distribution approximated by the new composite graph Laplacian $L$ in (\ref{equ:L}) for protein domain ranking,
we introduce a new multi-graph regularization term.
By substituting (\ref{equ:L}) into (\ref{equ:Og}), we get the augmented
objective function term in an enlarged parameter space as,

\begin{equation}
\label{equ:Omultig}
\begin{aligned}
\underset{\textbf{f},\mu}{min}~ &O^{multig}(\textbf{f},\mu)
=\sum_{m=1}^M \mu_m (\textbf{f}^\top L_m \textbf{f})\\
s.t.~&\sum_{m=1}^M \mu_m=1,~\mu_m\geq 0.
\end{aligned}
\end{equation}
where $\mu=[\mu_1,\cdots,\mu_M]^\top$ is the graph weight vector.

\subsubsection *{Off-line supervised multiple graph learning}

In the on-line querying procedure, the relevance of query $x_q$ to database protein domains is unknown and
thus the optimal graph weights $\mu$ cannot be learned in a supervised way.
However, all the SCOP-fold labels of protein domain in the database are known, making the supervised learning of $\mu$ in an off-line
way possible.
We treat each database protein domain $x_q\in \mathcal{D},~q=1,\cdots,N$ as a query in the off-line learning
and
all the items of its relevant vector $\textbf{y}_q=[{y}_{1q},\cdots,{y}_{Nq}]^\top $ as known because all the
SCOP-fold labels are known for all the database protein domains as,

\begin{equation}
\begin{aligned}
\textbf{y}_{iq}=
\left\{\begin{matrix}
1 &, if~l_i=l_q\\
0 &, else
\end{matrix}\right.
\end{aligned}
\end{equation}
Therefore, we set $U=I^{N\times N}$ as a $N\times N$ identity matrix.
The ranking score vector of the $q$-th database protein domain is also defined as $\textbf{f}_q=[y_{1q},\cdots,y_{Nq}]^\top$.
Substituting $\textbf{f}_q$, $\textbf{y}_q$ and $U$ to (\ref{equ:Qr}) and (\ref{equ:Omultig}) and combining them, we have the optimization
problem for the $q$-th database
protein domain as,

\begin{equation}
\label{equ:fq}
\begin{aligned}
\underset{\textbf{f}_q,\mu}{min}~ &O(\textbf{f}_q,\mu)
=(\textbf{f}_q-\textbf{y}_q)^\top (\textbf{f}_q-\textbf{y}_q)+\alpha \sum_{m=1}^M \mu_m (\textbf{f}_q^\top L_m \textbf{f}_q) + \beta ||\mu||^2\\
s.t.~&\sum_{m=1}^M \mu_m=1,~\mu_m\geq 0.
\end{aligned}
\end{equation}
To avoid the parameter $\mu$ over-fitting to
one single graph, we also introduce the $l_2$ norm
regularization term $||\mu||^2$ to the object function.
The difference between $f_q$ and $y_q$ should be noted:
$f_q \in \{1,0\}^N$ plays the role of the given ground truth in the supervised learning procedure,
while $y_q  \in
\mathbb{R}^N$ is the variable to be solved.
While $f_q$ is the ideal solution of $y_q$, it is not always achieved after the learning.
Thus, we introduce the first term in (\ref{equ:fq})to make $y_q$ as similar to $f_q$ as possible during the learning procedure.

\textbf{Object function}:
Using all protein domains in the database $q=1,\dots,N$ as queries to learn $\mu$, we obtain the final objective function of
supervised multiple graph weighting and protein domain ranking as,

\begin{equation}
\label{equ:OFmu}
\begin{aligned}
\underset{F,\mu}{min}~ O(F,\mu)
&=\sum_{q=1}^N \left[ (\textbf{f}_q-\textbf{y}_q)^\top (\textbf{f}_q-\textbf{y}_q)
+\alpha \sum_{m=1}^M \mu_m (\textbf{f}_q^\top L_m \textbf{f}_q) \right]+ \beta ||\mu||^2\\
&=Tr \left[ (F-Y)^\top (F-Y)\right]+\alpha \sum_{m=1}^M \mu_m Tr(F^\top L_m F) + \beta ||\mu||^2\\
s.t.~&\sum_{m=1}^M \mu_m=1,~\mu_m\geq 0.
\end{aligned}
\end{equation}
where $F=[\textbf{f}_1,\cdots,\textbf{f}_N]$ is the ranking score matrix with the $q$-th column as the ranking score vector of $q$-th
protein domain, and $Y=[\textbf{y}_1,\cdots,\textbf{y}_N]$ is the relevance matrix with the $q$-th column as the relevance vector of the $q$-th
protein domain.

\textbf{Optimization}:
Because direct optimization to (\ref{equ:OFmu}) is difficult, instead we
adopt an iterative, two-step strategy to alternately optimize
$F$ and $\mu$. At each iteration, either $F$ or $\mu$
is optimized while the other is fixed, and then the
roles are switched. Iterations are repeated
until
a maximum number of iterations is
reached.

\begin{itemize}
\item \emph{Optimizing $F$}:
By fixing $\mu$, the analytic solution for
 (\ref{equ:OFmu}) can be easily obtained by setting the
derivative of $O(F,\mu)$ with respect to $F$ to zero. That is,

\begin{equation}
\label{equ:F}
\begin{aligned}
\frac{\partial O(F,\mu)}{\partial F}&=2  (F-Y) + 2 \alpha \sum_{m=1}^M \mu_m (L_m F)=0\\
F&=( I+ \alpha \sum_{m=1}^M \mu_m L_m)^{-1}  Y
\end{aligned}
\end{equation}

\item \emph{Optimizing $\mu$}:
By fixing $F$ and removing items irrelevant to $\mu$ from (\ref{equ:OFmu}),
the optimization
problem (\ref{equ:OFmu}) is reduced to,

\begin{equation}
\label{equ:LP}
\begin{aligned}
\underset{\mu}{min}~
&\alpha\sum_{m=1}^M \mu_m Tr (F^\top L_m F)+\beta ||\mu||^2\\
&=\alpha\sum_{m=1}^M \mu_m e_m + \beta \sum_{m=1}^M \mu^2\\
&=\alpha e^\top \mu  + \beta \mu^\top \mu\\
s.t.~&\sum_{m=1}^M \mu_m=1,~\mu_m\geq 0.
\end{aligned}
\end{equation}
where $e_m=Tr (F^\top L_m F)$ and $e=[e_1,\cdots,e_M]^\top$.
The
optimization of (\ref{equ:LP}) with respect to the graph weight $\mu$
can then be solved as a standard quadratic programming (QP) problem \cite{Stivala2009}.
\end{itemize}

\textbf{Off-line algorithm}: The off-line $\mu$ learning algorithm is summarized as Algorithm \ref{alg:offline}.

\begin{algorithm}[h!]
\caption{MultiG-Rank: off-line graph weights learning algorithm.}
\label{alg:offline}
\begin{algorithmic}
\REQUIRE Candidate graph Laplacians set $\mathcal{T}$;
\REQUIRE SCOP type label set of database protein domains $\mathcal{L}$;
\REQUIRE Maximum iteration number $T$;
\STATE Construct the relevance matrix $Y=[y_{iq}]^{N\times N}$ where $y_{iq}$ if $l_i=l_q$, 0 otherwise;
\STATE Initialize the graph weights as $\mu_m^0=\frac{1}{M}$, $m=1,\cdots,M$;

\FOR{$t=1,\cdots,T$}
\STATE Update the ranking score matrix $F^t$ according to previous $\mu_m^{t-1}$ by (\ref{equ:F});
\STATE Update the graph weight $\mu^t$ according to updated $F^t$ by (\ref{equ:LP});
\ENDFOR

\STATE Output graph weight $\mu=\mu^{t}$.

\end{algorithmic}
\end{algorithm}

\subsubsection *{On-line ranking regularized by multiple graphs}
Given a newly discovered protein domain submitted by a user as query $x_0$,
its SCOP type label $l_0$ will be unknown and the domain will not be in the database $\mathcal{D}=\{x_1,\cdots,x_N\}$.
To compute the ranking scores of $x_i\in \mathcal{D}$ to query $x_0$, we
extend the size of database to $N + 1$  by adding $x_0$ into the database and
then solve the ranking score vector
for $x_0$ which is defined as $\textbf{f} =[f_0,\cdots,f_N] \in \mathbb{R}^{N+1}$ using (\ref{equ:Of}).
The parameters in (\ref{equ:Of}) are constructed as follows:

\begin{itemize}
\item \textbf{Laplacian matrix $L$}:
We first compute the $m$ graph weight
matrices $\{W_m\}_{m=1}^{M}\in \mathbb{R}^{(N+1)\times(N+1)}$ with their corresponding Laplacian matrices
$\{L_m\}_{m=1}^{M}\in \mathbb{R}^{(N+1)\times(N+1)}$ for the extended database $\{x_0,x_1,\cdots,x_N\}$.
Then with the graph weight $\mu$ learned by Algorithm \ref{alg:offline}, the new Laplacian matrix $L$
can be computed as in (\ref{equ:L}).

\emph{On-line graph weight computation}:
When a new query $x_0$ is added to the database, we calculate its $K$ nearest
neighbors in the database $\mathcal{D}$ and the corresponding weights $W_{0j}$ and $W_{j0},j=1,\cdots,N$.
If adding this new query to the database does not affect the
graph i n the database space, the neighbors and weights $W_{ij},i,j=1,\cdots,N$
for the protein domains in the database are fixed and can be pre-computed off-line.
Thus, we only need to compute $N$ edge weights for each graph instead of $(N+1)\times (N+1)$.

\item \textbf{Relevance vector $y$}:
The relevance vector for $x_0$ is defined as $\textbf{y} =[y_0,\cdots,y_N]^\top \in \{1,0\}^{N+1} $ with only $y_{0}=1$ known and $y_i$,
$i=1,\cdots,N$ unknown.

\item \textbf{Matrix $U$}:
In this situation, $U$ is a $(N+1)\times (N+1)$ diagonal matrix with $U_{00}=1$ and $U_{ii}=0$, $i=1,\cdots,N$.
\end{itemize}
Then the ranking score vector $f$ can be solved as,

\begin{equation}
\label{equ:f}
\begin{aligned}
\textbf{f}=(U+\alpha L)^{-1} U \textbf{y}
\end{aligned}
\end{equation}
The on-line ranking algorithm is summarized as Algorithm \ref{alg:inline}.

\begin{algorithm}[h!]
\caption{MultiG-Rank: on-line ranking algorithm.}\label{alg:inline}
\begin{algorithmic}
\REQUIRE protein domain database $\mathcal{D}=\{x_1,\cdots,x_N\}$;
\REQUIRE Query protein domain $x_0$;
\REQUIRE Graph weight $\mu$;
\STATE Extend the database to $(N+1)$ size by adding $x_0$ and compute $M$
graph Laplacians of the extended database;
\STATE Obtain multiple graph Laplacian $L$ by linear combination of $M$ graph Laplacians with weight $\mu$ as in (\ref{equ:L});
\STATE Construct the relevance vector $\textbf{y}\in \mathbb{R}^{(N+1)}$
where $y_{0}=1$ and diagonal matrix $U\in \mathbb{R}^{(N+1)\times (N+1)}$ with $U_{ii}=1$ if $i=0$ and 0 otherwise;
\STATE Solve the ranking vector $\textbf{f}$ for $x_0$ as in (\ref{equ:f});
\STATE Ranking protein domains in $\mathcal{D}$ according to ranking scores $\textbf{f}$ in descending order.
\end{algorithmic}
\end{algorithm}

\subsection*{Protein domain database and query set}

We used the
SCOP 1.75A database \cite{ASTRAL2004} to construct the database and query set.
In the SCOP 1.75A database, there are
49,219 protein domain PDB entries and 135,643 domains,
belonging to 7 classes and 1,194 SCOP fold types.

\subsubsection*{Protein domain database}
Our protein domain database was selected from \emph{ASTRAL SCOP 1.75A} set \cite{ASTRAL2004}, a subset of the SCOP (Structural Classification of Proteins)1.75A database which was released in March 15, 2012 \cite{ASTRAL2004}.
 ASTRAL SCOP 1.75A 40\%) \cite{ASTRAL2004}, a genetic domain sequence subset,
was used  as our protein domain database $\mathcal{D}$.
This database was selected from SCOP 1.75A database so that the selected domains have less than 40\% identity to each other.
There are a total of 11,212 protein domains in the ASTRAL SCOP 1.75A 40\% database belonging to 1,196 SCOP fold types. The ASTRAL  database is available on-line at
http://scop.berkeley.edu.
The number of protein domains in each SCOP fold varies from 1 to 402.
The distribution of protein domains with the different fold types is shown in Fig. \ref{fig:FigProNum}.
Many previous studies evaluated ranking performances
using the older version of the ASTRAL SCOP dataset (ASTRAL SCOP 1.73 95\%) that was released in 2008 \cite{Zhang2010}.

\begin{figure}[h!]
\centering
\caption{Distribution of protein domains with different fold types in the ASTRAL SCOP 1.75A 40\% database.}
\label{fig:FigProNum}
\end{figure}

\subsubsection*{Query set}
We also randomly selected 540 protein domains from the SCOP 1.75A database to construct a query set.
For each query protein domain that we selected we ensured that there was at least one protein domain belonging to the same SCOP fold type
in the
ASTRAL SCOP 1.75A 40\% database, so that for each query, there was at least one "positive" sample in
the protein domain database.
However, it should be noted that the 540 protein domains in the query data set
were randomly selected and do not necessarily represent 540 different folds.
Here we call our query set the \emph{540 query} dataset because it contains 540 protein domains from the
SCOP 1.75A database.

\subsection*{Evaluation metrics}

A ranking procedure is run against the protein domains database using a query domain. A list of all matching protein domains
along with their ranking scores is returned.
 We adopted the same evaluation metric framework as was described previously \cite{Zhang2010}, and
used the receiver operating characteristic (ROC) curve,
the area under the ROC curve (AUC),
and the recall-precision curve
to evaluate the ranking accuracy.
Given a query protein domain $x_q$ belonging to the SCOP fold $l_q$, a list of
protein domains is returned from the database by the on-line MultiG-Rank algorithm or other ranking
methods. For a database protein domain $x_r$ in the returned list, if its fold label $l_r$ is the same as that of
$x_q$, i.e. $l_r = l_q$ it is identified as a true positive (TP), else it is identified as a false positive (FP).
For a database protein domain $x_{r'}$ not in the returned list, if its fold label $l_{r'}= l_q$, it will be identified
as a true negative (TN), else it is a false negative (FN). The true positive rate (TPR),
false positive rate (FPR), recall, and precision can then be computed based on the above statistics as follows:

\begin{equation}
\begin{aligned}
&TPR =\frac{ TP}{ TP+ FN},~&FPR= \frac{ FP}{ FP+ TN}\\
&recall=\frac{ TP}{ TP+ FN},~&precision =\frac{TP}{ TP+ FP}
\end{aligned}
\end{equation}
By varying the length of the returned list, different $TPR$, $FRP$, recall and precision values are obtained.
\begin{description}
\item[ROC curve] Using $FPR$ as the abscissa and $TPR$ as the ordinate, the ROC curve can be plotted. For a
high-performance ranking system, the ROC curve should be as close to the top-left corner as possible.
\item[Recall-precision curve]
Using recall as the abscissa and precision as the ordinate, the recall-precision curve
can be plotted. For a high-performance ranking system, this curve should be close to the top-right
corner of the plot.
\item[AUC]
The AUC is computed as a single-figure measurement of the quality of an ROC curve. AUC is averaged over all the queries to evaluate the performances of different ranking methods.
\end{description}

\section*{Results}

We first compared our MultiG-Rank against
several popular graph-based ranking score learning methods for ranking protein {domains}.
We then evaluated the ranking performance of MultiG-Ranking against other
protein domain ranking methods using different protein domain comparison strategies.
Finally, a case study of a TIM barrel fold is described.

\subsubsection*{Comparison of MultiG-Rank against other graph-based ranking methods}

We compared our MultiG-Rank to two graph-based ranking methods, G-Rank and GT \cite{BaiGT2010}, and
against the pairwise protein domain comparison based ranking method proposed in \cite{Zhang2010} as a  baseline method (Fig. \ref{fig:GraphROC}) .
 The evaluations were conducted with the 540 query domains form the \emph{540 query} set. The average
ranking performance was computed over these 540 query runs.

\begin{figure}[h!]
\centering
\caption{
Comparison of MultiG-Rank against other protein domain ranking methods.
Each curve represents a graph-based ranking score learning algorithm.
MultiG-Rank, the Multiple Graph regularized Ranking algorithm; G-Rank, Graph regularized Ranking; GT, graph transduction; Pairwise Rank, pairwise protein domain ranking method \cite{Zhang2010}
(a) ROC curves of the different ranking methods;
(b) Recall-precision curves of the different ranking methods.
}\label{fig:GraphROC}
\end{figure}

The figure shows the ROC and the recall-precision curves obtained using the different graph ranking
methods.
 As can be seen, the MultiG-Rank algorithm
significantly outperformed the other graph-based ranking algorithms;
the precision difference got larger
as the recall value increased and then tend to converge as the precision tended towards zero (Fig. \ref{fig:GraphROC} (b)).
The G-Rank
algorithm outperformed GT in most cases;
however, both G-Rank and GT were much better than the
pairwise ranking which neglects the global distribution of the protein domain database.

The AUC results for the
different ranking methods on the \emph{540 query} set are tabulated in Table \ref{tab:AUCgraph}.
As shown,
the proposed MultiG-Rank consistently outperformed the
other three methods on the 540 query set against our protein domain
database, achieving a gain in AUC of  0.0155,  0.0210
and  0.0252 compared
with G-Rank, GT and Pairwise Rank, respectively.
Thus, we have shown that the ranking precision can be improved significantly using our algorithm.

We have made three observations from the results
listed in Table \ref{tab:AUCgraph}:
\begin{enumerate}
\item
G-Rank and GT produced similar performances
on our protein domain database,
indicating that there is no significant difference
in the performance of the graph \emph{transduction} based or
graph \emph{regularization based} single graph ranking methods for unsupervised
learning of the ranking scores.
\item
Pairwise ranking produced the worst performance even though the method uses a carefully selected
similarity function as reported in \cite{Zhang2010}.
One reason for the poorer performance is that similarity computed
by pairwise ranking is
focused on detecting statistically significant pairwise differences only, while more subtle sequence similarities are missed.
Hence, the variance among different fold types cannot be accurately estimated when the global distribution is neglected and only the protein domain pairs are considered.
Another possible reason is that pairwise ranking usually
produces a better performance when there is only a small number of
protein domains in the database; therefore, because our database  contains a large
number of protein domains, the ranking
performance of the pairwise ranking method was poor.
\item
MultiG-Rank produced the best ranking performance, implying that both
the discriminant and geometrical information in the protein domain database are important for accurate ranking.
In MultiG-Rank, the geometrical information is estimated by multiple graphs and the discriminant information
is included by using the SCOP-fold type labels to learn the graph weights.
 \end{enumerate}

\subsubsection*{Comparison of MultiG-Rank with other protein domain ranking methods}

We compare the MultiG-Rank against several other popular protein domain ranking methods:
IR Tableau \cite{Zhang2010}, QP tableau \cite{Stivala2009}, YAKUSA \cite{YAKUSA2005}, and
SHEBA\cite{SHEBA1990}. For the query domains and the protein domain database we used the \emph{540 query} set and  the ASTRAL SCOP 1.75A 40\% database, respectively.
The YAKUSA software source code was downloaded from http://wwwabi.snv.jussieu.fr/YAKUSA , compiled and used for ranking.
We used the ``make Bank" shell script (http://wwwabi.snv.jussieu.fr/YAKUSA) which calls the phipsi program (Version 0.99 ABI, June  1993) to format the database.  YAKUSA compares a query domain to a database and returns a list of the protein domains along with ranks and ranking scores.
We used the default parameters of YAKUSA to perform the ranking of the protein domains in our database.
The SHEBA software (version 3.11) source code was downloaded from https://ccrod.cancer.gov/confluence/display/CCRLEE/SHEBA,
complied and used it for ranking.
The protein domain database was converted to ``.env" format and the
pairwise alignment was performed between each query domain and
each database domain to obtain the alignment scores.
 First, we compared the different
protein domain-protein domain ranking methods and computed their similarity or dissimilarity.
 An ordering technique was devised to detect
hits by taking the similarities
between data pairs as input.
For our MultiG-Rank, the ranking score was used as a measure of protein domain-protein domain similarly.
The
ranking results were evaluated based on
the ROC and recall-precision curves as shown in Fig. \ref{fig:RankingROC}.
The AUC values are
given in Table \ref{tab:AUCranking}.

\begin{figure}[h!]
\centering
\caption{
Comparison of the performances of protein domain ranking algorithms.
(a) ROC curves for
different field-specific protein domain ranking algorithms.
TPR, true positive rate; FPR, false positive rate.
(b) Recall-precision curves for different field-specific protein domain ranking algorithms.}
\label{fig:RankingROC}
\end{figure}

The results in Table \ref{tab:AUCranking} show that with the
advantage of exploring data characteristics from various
graphs, MultiG-Rank can achieve significant improvements
in the ranking outcomes; in particular, AUC is increased from 0.9478
to 0.9730 in MultiG-Rank  which uses the same Tableau feature as IR Tableau.
MultiG-Rank also outperforms
QP Tableau, SHEBA, and YAKUSA; and AUC improves from
0.9364, 0.9421 and 0.9537, respectively, to 0.9730 with MultiG-Rank.
Furthermore, because of its better use of effective
protein domain descriptors, IR Tableau outperforms QP Tableau.

To evaluate the effect of using protein domain descriptors for ranking instead of
direct protein domain structure comparisons, we compared IR Tableau with YAKUSA
and SHEBA. The main differences between them are
that IR Tableau considers both protein domain feature extraction and comparison
procedures, while YAKUSA
and SHEBA compare only pairs of protein domains directly.
The quantitative results in Table \ref{tab:AUCranking} show
that, even by using the additional information from the
protein domain descriptor, IR Tableau  does not outperform YAKUSA
.
 This result strongly suggests that ranking performance improvements are achieved mainly by
graph regularization and not by using the power of a protein domain descriptor.

Plots of TPR versus
FPR obtained using MultiG-Rank and various field-specific protein domain ranking methods as
the ranking algorithms are shown in Fig. \ref{fig:RankingROC} (a) and the recall-precision curves obtained using them are shown in Fig. \ref{fig:RankingROC} (b).  As can be seen from the figure, in most cases, our MultiG-Rank algorithm
significantly outperforms the other protein domain ranking algorithms.
The performance differences get larger
as the length of the returned protein domain list increases.
The YAKUSA
algorithm outperforms SHEBA, IR Tableau and QP Tableau  in most cases.
When only a few protein domains are returned to the query,
the sizes of both the true positive samples and the false positive samples are small, showing that, in this case, all the algorithms yield
low FPR and TPR.
As the number of returned protein domains
increases, the TPR of all of the algorithms
increases. However, MultiG-Rank tends to converge
when the FPR is more than 0.3,
whereas the other ranking algorithms seems
to converge only when the FPR is more than 0.5.

\subsubsection*{Case Study of the TIM barrel fold}
Besides considering the results obtained for the whole database, we also
studied an important protein fold, the TIM beta/alpha-barrel fold (c.1).
The TIM barrel is a conserved protein fold that consists of eight $\alpha$-helices and
eight parallel $\beta$-strands that alternate along the peptide backbone \cite{TMI2010}.
TIM barrels are one of the most common protein folds.
In the ASTRAL SCOP 1.75A \%40 database, there are a total of 373 proteins belonging to
33 different superfamilies and 114 families that have TIM beta/alpha-barrel SCOP fold type domains,.
In this case study, the TIM beta/alpha-barrel domains from the query set
were used to rank
all the protein domains in the database.
The ranking was evaluated both at the fold level of the SCOP classification
and at lower levels of the SCOP classification (ie. superfamily level and family level).
To evaluate the ranking performance, we defined "true positives"
at three levels:
\begin{description}
\item[Fold level] When the returned database protein domain is from the same fold type as the query protein domain.
\item[Superfamily level] When the returned database protein domain is from the same superfamily as the query protein domain.
\item[Family level] When the returned database protein domain is from the same family as the query protein domain.
\end{description}

The ROC and the recall-precision plots of the protein domain ranking results of MultiG-Rank
for the query TIM beta/alpha-barrel domain
at the three levels are
given in Fig. \ref{fig:ResultsTMI}.
The graphs were learned using the labels at the family, superfamily and the fold level.
The results show that
the ranking performance at the fold level is better than at the other two levels; however, although
the
performances at the lower levels, superfamily
and family, are not superior to that at the fold level,
they are still good.
One important factor is that when the relevance at the lower levels was measured, a much fewer number of protein domains
in the database were relevant to the queries,
making it more difficult to retrieve the relevant protein domains precisely.
For example, a query belonging to the
family of phosphoenolpyruvate mutase/Isocitrate lyase-like (c.1.12.7) matched 373 database protein domains at the fold level because this family
has 373 protein domains in the ASTRAL SCOP 1.75A \%40 database.
On the other hand, only 14 and four protein domains were relevant to the query at the superfamily and family levels respectively.

\begin{figure}[h!]
\centering
\caption{
{
Ranking results for the case study using the TIM beta/alpha-barrel domain as the query.
(a)
ROC curves of the ranking results for the TIM beta/alpha-barrel domain at the
fold, superfamily, and family levels.
TPR, true positive rate; FPR, false positive rate.
(b)
Recall-precision curves of the ranking results for the TIM beta/alpha-barrel domain
at the fold, superfamily, and family levels.}
}
\label{fig:ResultsTMI}
\end{figure}

\section*{Conclusion}
\label{sec:conclu}

The proposed MultiG-Rank method introduces a new paradigm to
fortify the broad scope of existing graph-based ranking
techniques. The main advantage of MultiG-Rank lies in its ability
to represent the learning of a unified space of ranking scores for protein domain database
in multiple graphs.
Such flexibility
is important in tackling complicated protein domain ranking problems
because it allows more prior knowledge to be explored for
effectively analyzing a given protein domain database, including the possibility of choosing
a proper set of graphs to better characterize
diverse databases, and the ability to adopt a multiple graph-based ranking method to appropriately
model relationships among the protein domains.
Here, MultiG-Rank has been
evaluated comprehensively on a carefully selected subset of the ASTRAL SCOP 1.75 A protein domain database.
The promising experimental results that were obtained further
confirm the usefulness of our ranking score learning approach.

\bigskip

\section*{Competing interests}

The authors declare no competing interests.

\section*{Author's contributions}
JW invented the algorithm, performed the experiments and
drafted the manuscript.
HB drafted the manuscript.
XG
supervised the study and made critical changes to  the manuscript.
All the authors have
approved the final manuscript.

\section*{Acknowledgements}
\ifthenelse{\boolean{publ}}{\small}{}
The study was supported by grants from National Key Laboratory for Novel
Software Technology, China (Grant No. KFKT2012B17), 2011 Qatar Annual
Research Forum Award (Grant No. ARF2011), and King Abdullah University of
Science and Technology (KAUST), Saudi Arabia. We appreciate the valuable
comments from Prof. Yuexiang Shi, Xiangtan University, China.




\newcommand{\BMCxmlcomment}[1]{}

\BMCxmlcomment{

<refgrp>

<bibl id="B1">
  <title><p>{HMM-FRAME: accurate protein domain classification for metagenomic
  sequences containing frameshift errors}</p></title>
  <aug>
    <au><snm>Zhang</snm><fnm>Y</fnm></au>
    <au><snm>Sun</snm><fnm>Y</fnm></au>
  </aug>
  <source>{BMC Bioinformatics}</source>
  <pubdate>{2011}</pubdate>
  <volume>{12}</volume>
  <issue>{198}</issue>
</bibl>

<bibl id="B2">
  <title><p>{Using context to improve protein domain
  identification}</p></title>
  <aug>
    <au><snm>Ochoa</snm><fnm>A</fnm></au>
    <au><snm>Llinas</snm><fnm>M</fnm></au>
    <au><snm>Singh</snm><fnm>M</fnm></au>
  </aug>
  <source>{BMC Bioinformatics}</source>
  <pubdate>{2011}</pubdate>
  <volume>{12}</volume>
  <issue>{90}</issue>
</bibl>

<bibl id="B3">
  <title><p>{A fast indexing approach for protein structure
  comparison}</p></title>
  <aug>
    <au><snm>Zhang</snm><fnm>L</fnm></au>
    <au><snm>Bailey</snm><fnm>J</fnm></au>
    <au><snm>Konagurthu</snm><fnm>AS</fnm></au>
    <au><snm>Ramamohanarao</snm><fnm>K</fnm></au>
  </aug>
  <source>{BMC Bioinformatics}</source>
  <pubdate>{2010}</pubdate>
  <volume>{11}</volume>
  <issue>{1}</issue>
</bibl>

<bibl id="B4">
  <title><p>{Tableau-based protein substructure search using quadratic
  programming}</p></title>
  <aug>
    <au><snm>Stivala</snm><fnm>A</fnm></au>
    <au><snm>Wirth</snm><fnm>A</fnm></au>
    <au><snm>Stuckey</snm><fnm>PJ</fnm></au>
  </aug>
  <source>{BMC Bioinformatics}</source>
  <pubdate>{2009}</pubdate>
  <volume>{10}</volume>
  <fpage>{153}</fpage>
</bibl>

<bibl id="B5">
  <title><p>{Fast and accurate protein substructure searching with simulated
  annealing and GPUs}</p></title>
  <aug>
    <au><snm>Stivala</snm><fnm>AD</fnm></au>
    <au><snm>Stuckey</snm><fnm>PJ</fnm></au>
    <au><snm>Wirth</snm><fnm>AI</fnm></au>
  </aug>
  <source>{BMC Bioinformatics}</source>
  <pubdate>{2010}</pubdate>
  <volume>{11}</volume>
  <issue>{446}</issue>
</bibl>

<bibl id="B6">
  <title><p>{Learning Context-Sensitive Shape Similarity by Graph
  Transduction}</p></title>
  <aug>
    <au><snm>Bai</snm><fnm>X</fnm></au>
    <au><snm>Yang</snm><fnm>X</fnm></au>
    <au><snm>Latecki</snm><fnm>LJ</fnm></au>
    <au><snm>Liu</snm><fnm>W</fnm></au>
    <au><snm>Tu</snm><fnm>Z</fnm></au>
  </aug>
  <source>{IEEE Transactions on Pattern Analysis and Machine
  Intelligence}</source>
  <pubdate>{2010}</pubdate>
  <volume>{32}</volume>
  <issue>{5}</issue>
  <fpage>{861</fpage>
  <lpage>874}</lpage>
</bibl>

<bibl id="B7">
  <title><p>{YAKUSA: A fast structural database scanning method}</p></title>
  <aug>
    <au><snm>Carpentier</snm><fnm>M</fnm></au>
    <au><snm>Brouillet</snm><fnm>S</fnm></au>
    <au><snm>Pothier</snm><fnm>J</fnm></au>
  </aug>
  <source>{Proteins-Structure Function and Bioinformatics}</source>
  <pubdate>{2005}</pubdate>
  <volume>{61}</volume>
  <issue>{1}</issue>
  <fpage>{137</fpage>
  <lpage>151}</lpage>
</bibl>

<bibl id="B8">
  <title><p>{Protein structure alignment using environmental
  profiles}</p></title>
  <aug>
    <au><snm>Jung</snm><fnm>J</fnm></au>
    <au><snm>Lee</snm><fnm>B</fnm></au>
  </aug>
  <source>{Protein Engineering}</source>
  <pubdate>{2000}</pubdate>
  <volume>{13}</volume>
  <issue>{8}</issue>
  <fpage>{535</fpage>
  <lpage>543}</lpage>
</bibl>

<bibl id="B9">
  <title><p>{Protein comparison at the domain architecture level}</p></title>
  <aug>
    <au><snm>Lee</snm><fnm>B</fnm></au>
    <au><snm>Lee</snm><fnm>D</fnm></au>
  </aug>
  <source>{BMC Bioinformatics}</source>
  <pubdate>{2009}</pubdate>
  <volume>{10}</volume>
</bibl>

<bibl id="B10">
  <title><p>{Protein ranking by semi-supervised network
  propagation}</p></title>
  <aug>
    <au><snm>Weston</snm><fnm>J</fnm></au>
    <au><snm>Kuang</snm><fnm>R</fnm></au>
    <au><snm>Leslie</snm><fnm>C</fnm></au>
    <au><snm>Noble</snm><fnm>WS</fnm></au>
  </aug>
  <source>{BMC Bioinformatics}</source>
  <pubdate>{2006}</pubdate>
  <volume>{7}</volume>
  <issue>{1}</issue>
</bibl>

<bibl id="B11">
  <title><p>{Graph Regularized Nonnegative Matrix Factorization for Data
  Representation}</p></title>
  <aug>
    <au><snm>Cai</snm><fnm>D</fnm></au>
    <au><snm>He</snm><fnm>X</fnm></au>
    <au><snm>Han</snm><fnm>J</fnm></au>
    <au><snm>Huang</snm><fnm>TS</fnm></au>
  </aug>
  <source>{IEEE Transactions on Pattern Analysis and Machine
  Intelligence}</source>
  <pubdate>{2011}</pubdate>
  <volume>{33}</volume>
  <issue>{8}</issue>
  <fpage>{1548</fpage>
  <lpage>1560}</lpage>
</bibl>

<bibl id="B12">
  <title><p>{Bias in error estimation when using cross-validation for model
  selection}</p></title>
  <aug>
    <au><snm>Varma</snm><fnm>S</fnm></au>
    <au><snm>Simon</snm><fnm>R</fnm></au>
  </aug>
  <source>{BMC Bioinformatics}</source>
  <pubdate>{2006}</pubdate>
  <volume>{7}</volume>
  <issue>{91}</issue>
</bibl>

<bibl id="B13">
  <title><p>{Annotation of protein residues based on a literature analysis:
  cross-validation against UniProtKb}</p></title>
  <aug>
    <au><snm>Nagel</snm><fnm>K</fnm></au>
    <au><snm>Jimeno Yepes</snm><fnm>A</fnm></au>
    <au><snm>Rebholz Schuhmann</snm><fnm>D</fnm></au>
  </aug>
  <source>{BMC Bioinformatics}</source>
  <pubdate>{2009}</pubdate>
  <volume>{10}</volume>
</bibl>

<bibl id="B14">
  <title><p>Ensemble Manifold Regularization</p></title>
  <aug>
    <au><snm>Geng</snm><fnm>B.</fnm></au>
    <au><snm>Tao</snm><fnm>D.</fnm></au>
    <au><snm>Xu</snm><fnm>C.</fnm></au>
    <au><snm>Yang</snm><fnm>L.</fnm></au>
    <au><snm>Hua</snm><fnm>X.</fnm></au>
  </aug>
  <source>IEEE Transactions on Pattern Analysis and Machine
  Intelligence</source>
  <pubdate>2012</pubdate>
  <volume>PP</volume>
  <issue>99</issue>
  <fpage>1</fpage>
</bibl>

<bibl id="B15">
  <title><p>{A semi-supervised learning approach to predict synthetic genetic
  interactions by combining functional and topological properties of functional
  gene network}</p></title>
  <aug>
    <au><snm>You</snm><fnm>ZH</fnm></au>
    <au><snm>Yin</snm><fnm>Z</fnm></au>
    <au><snm>Han</snm><fnm>K</fnm></au>
    <au><snm>Huang</snm><fnm>DS</fnm></au>
    <au><snm>Zhou</snm><fnm>X</fnm></au>
  </aug>
  <source>{BMC Bioinformatics}</source>
  <pubdate>{2010}</pubdate>
  <volume>{11}</volume>
  <issue>{343}</issue>
</bibl>

<bibl id="B16">
  <title><p>{Combining active learning and semi-supervised learning techniques
  to extract protein interaction sentences}</p></title>
  <aug>
    <au><cnm>{Min Song}</cnm></au>
    <au><cnm>{Hwanjo Yu}</cnm></au>
    <au><cnm>{Wook-Shin Han}</cnm></au>
  </aug>
  <source>{BMC Bioinformatics}</source>
  <pubdate>2011</pubdate>
  <fpage>{S4(11pp.)}</fpage>
</bibl>

<bibl id="B17">
  <title><p>{BLProt: prediction of bioluminescent proteins based on support
  vector machine and relieff feature selection}</p></title>
  <aug>
    <au><snm>Kandaswamy</snm><fnm>KK</fnm></au>
    <au><snm>Pugalenthi</snm><fnm>G</fnm></au>
    <au><snm>Hazrati</snm><fnm>MK</fnm></au>
    <au><snm>Kalies</snm><fnm>KU</fnm></au>
    <au><snm>Martinetz</snm><fnm>T</fnm></au>
  </aug>
  <source>{BMC Bioinformatics}</source>
  <pubdate>{2011}</pubdate>
  <volume>{12}</volume>
  <issue>{345}</issue>
</bibl>

<bibl id="B18">
  <title><p>{POPISK: T-cell reactivity prediction using support vector machines
  and string kernels}</p></title>
  <aug>
    <au><snm>Tung</snm><fnm>CW</fnm></au>
    <au><snm>Ziehm</snm><fnm>M</fnm></au>
    <au><snm>Kaemper</snm><fnm>A</fnm></au>
    <au><snm>Kohlbacher</snm><fnm>O</fnm></au>
    <au><snm>Ho</snm><fnm>SY</fnm></au>
  </aug>
  <source>{BMC Bioinformatics}</source>
  <pubdate>{2011}</pubdate>
  <volume>{12}</volume>
  <issue>{446}</issue>
</bibl>

<bibl id="B19">
  <title><p>{Fast SCOP classification of structural class and fold using
  secondary structure mining in distance matrix}</p></title>
  <aug>
    <au><snm>Shi</snm><fnm>JY</fnm></au>
    <au><snm>Zhang</snm><fnm>YN</fnm></au>
  </aug>
  <source>{Pattern Recognition in Bioinformatics. Proceedings 4th IAPR
  International Conference, PRIB 2009}</source>
  <pubdate>2009</pubdate>
  <fpage>{344</fpage>
  <lpage>53}</lpage>
</bibl>

<bibl id="B20">
  <title><p>{Correcting Jaccard and other similarity indices for chance
  agreement in cluster analysis}</p></title>
  <aug>
    <au><snm>Albatineh</snm><fnm>AN</fnm></au>
    <au><snm>Niewiadomska Bugaj</snm><fnm>M</fnm></au>
  </aug>
  <source>{Advances in Data Analysis and Classification}</source>
  <pubdate>{2011}</pubdate>
  <volume>{5}</volume>
  <issue>{3}</issue>
  <fpage>{179</fpage>
  <lpage>200}</lpage>
</bibl>

<bibl id="B21">
  <title><p>{The ASTRAL Compendium in 2004}</p></title>
  <aug>
    <au><snm>Chandonia</snm><fnm>JM</fnm></au>
    <au><snm>Hon</snm><fnm>G</fnm></au>
    <au><snm>Walker</snm><fnm>NS</fnm></au>
    <au><snm>Lo Conte</snm><fnm>L</fnm></au>
    <au><snm>Koehl</snm><fnm>P</fnm></au>
    <au><snm>Levitt</snm><fnm>M</fnm></au>
    <au><snm>Brenner</snm><fnm>SE</fnm></au>
  </aug>
  <source>{Nucleic Acids Research}</source>
  <pubdate>{2004}</pubdate>
  <volume>{32}</volume>
  <issue>{SI}</issue>
  <fpage>{D189</fpage>
  <lpage>D192}</lpage>
</bibl>

<bibl id="B22">
  <title><p>{Detecting internally symmetric protein structures}</p></title>
  <aug>
    <au><snm>Kim</snm><fnm>C</fnm></au>
    <au><snm>Basner</snm><fnm>J</fnm></au>
    <au><snm>Lee</snm><fnm>B</fnm></au>
  </aug>
  <source>{BMC Bioinformatics}</source>
  <pubdate>{2010}</pubdate>
  <volume>{11}</volume>
  <issue>{303}</issue>
</bibl>

</refgrp>
} 

\newpage

\section*{Figures}
\subsection*{Figure 1 -
Distribution of protein domains with different fold types in the ASTRAL SCOP 1.75A 40\% database.}

~

\subsection*{Figure 2 - Comparison of MultiG-Rank against other protein domain ranking methods.}

Each curve represents a graph-based ranking score learning algorithm.
MultiG-Rank, the Multiple Graph regularized Ranking algorithm; G-Rank, Graph regularized Ranking; GT, graph transduction; Pairwise Rank, pairwise protein domain ranking method \cite{Zhang2010}
(a) ROC curves of the different ranking methods;
(b) Recall-precision curves of the different ranking methods.

%

\subsection*{Figure 3 -
Comparison of the performances of protein domain ranking algorithms.
}

(a) ROC curves for
different field-specific protein domain ranking algorithms.
TPR, true positive rate; FPR, false positive rate.
(b) Recall-precision curves for different field-specific protein domain ranking algorithms.

\subsection*{Figure 4 - Ranking results for the case study using the TIM beta/alpha-barrel domain as the query.
}

(a)
ROC curves of the ranking results for the TIM beta/alpha-barrel domain at the
fold, superfamily, and family levels.
TPR, true positive rate; FPR, false positive rate.
(b)
Recall-precision curves of the ranking results for the TIM beta/alpha-barrel domain
at the fold, superfamily, and family levels.

\newpage

\section*{Tables}

\subsection*{Table 1 - AUC results off different graph-based ranking methods.}

\begin{table}[h!]
\centering
\caption{AUC results off different graph-based ranking methods.}
\label{tab:AUCgraph}
\begin{tabular}{lc}
\hline
Method & AUC \\\hline\hline
MultiG-Rank & 0.9730 \\\hline
G-Rank & 0.9575 \\\hline
GT &0.9520\\\hline
Pairwise-Rank & 0.9478\\\hline
\end{tabular}
\end{table}

\subsection*{Table 2 - AUC results for different protein domain ranking methods.}

\begin{table}[h!]
\centering
\caption{AUC results for different protein domain ranking methods.}
\label{tab:AUCranking}
\begin{tabular}{lc}
\hline
Method & AUC \\\hline\hline
MultiG-Rank & 0.9730 \\\hline
IR Tableau & 0.9478 \\\hline
YAKUSA& 0.9537\\\hline
SHEBA& 0.9421\\\hline
QP tableau& 0.9364\\\hline
\end{tabular}
\end{table}

\end{bmcformat}
\end{document}